\documentclass[letterpaper, 10 pt, conference]{ieeeconf}  % Comment this line out if you need a4paper

\IEEEoverridecommandlockouts                              % This command is only needed if 
\usepackage{lmodern}
\usepackage[T1]{fontenc}

\usepackage{graphics} % for pdf, bitmapped graphics files
\usepackage{subcaption}
\usepackage{epsfig} % for postscript graphics files
\usepackage{mathptmx} % assumes new font selection scheme installed
\usepackage{times} % assumes new font selection scheme installed
\usepackage{amsmath} % assumes amsmath package installed
\usepackage{amssymb}  % assumes amsmath package installed
\usepackage{comment}
\usepackage[T1]{fontenc}
\usepackage{xcolor}
\usepackage{cleveref}
\usepackage{tabularx}
\usepackage{booktabs}

\usepackage{multirow}

\newcommand{\figref}[1]{Fig.~\ref{#1}}

\newcommand{\ourname}{PGR }

\title{\LARGE \bf
Can We Remove the Ground? Obstacle-aware Point Cloud\\ Compression for Remote Object Detection
}

\author{Pengxi Zeng $^{1}$, Alberto Presta$^{2*}$,  Jonah Reinis$^{3}$, \\ Dinesh Bharadia $^{1}$, Hang Qiu$^{4}$, and Pamela Cosman$^{1}$ % <-this % stops a space
\thanks{$^{1}$Dept. of Electrical and Computer Engineering, UC San Diego, CA, USA} % \emph{ \{p2zeng,dineshb,pcosman\}@ucsd.edu}}%
\thanks{$^{2}$Computer Science Department, University of Turin, Italy}%
\thanks{$^{3}$Case Western Reserve University, OH, USA}%
\thanks{$^{4}$ECE / CSE Dept., University of California Riverside, CA, USA}%
\thanks{$^{*}$Corresponding author, mail : alberto.presta@unito.it}%
}

\begin{document}

\maketitle
\thispagestyle{empty}
\pagestyle{empty}

%%%%%%%%%%%%%%%%%%%%%%%%%%%%%%%%%%%%%%%%%%%%%%%%%%%%%%%%%%%%%%%%%%%%%%%%%%%%%%%%
\begin{abstract}
Efficient point cloud (PC)  compression is crucial for streaming applications, such as augmented reality and cooperative perception.  Classic PC compression techniques encode all the points in a frame.
Tailoring compression towards perception tasks at the receiver side, 
we ask the question, "Can we remove the ground points during transmission without sacrificing the detection performance?"  Our study reveals a strong dependency on the ground from state-of-the-art (SOTA) 3D object detection models, especially on those points below and around the object. In this work, we propose a lightweight obstacle-aware Pillar-based Ground Removal (PGR) algorithm. \ourname filters out ground points that do not provide context to object recognition, significantly improving compression ratio without sacrificing the receiver side perception performance.
Not using heavy object detection or semantic segmentation models, \ourname is light-weight, highly parallelizable, and effective.
Our evaluations on KITTI and Waymo Open Dataset show that SOTA detection models work equally well with PGR removing 20-30\% of the points, with a speeding of 86 FPS. 
%Paired with SOTA compression algorithms, PGR can achieve up to XXXX\% transmission size reduction.
%PGR runs at 86 FPS, and it is robust to uneven road surfaces, sloped streets, and various object sizes.
\end{abstract}

%%%%%%%%%%%%%%%%%%%%%%%%%%%%%%%%%%%%%%%%%%%%%%%%%%%%%%%%%%%%%%%%%%%%%%%%%%%%%%%%

\begin{figure*}[h!]
    \centering
    \includegraphics[width=0.6\textwidth]{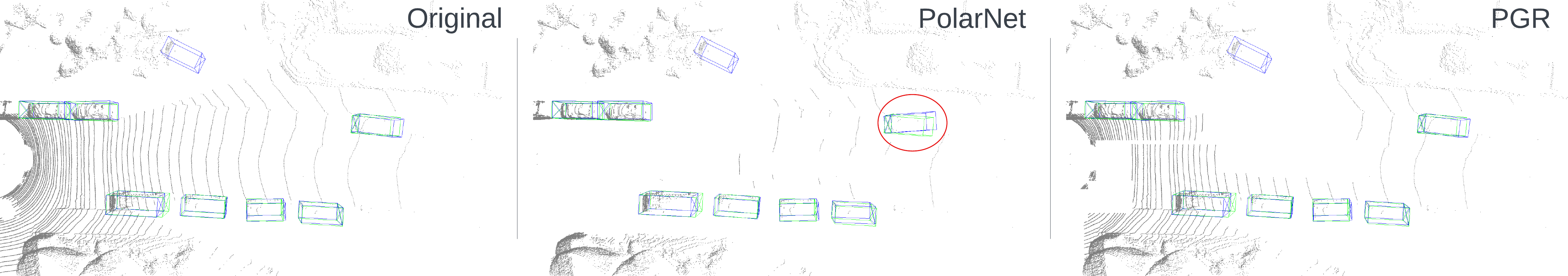}
    \caption{Object ground truth (green) and detection results (blue) using PVT-SSD with an input of original (left) PC, PC with semantic ground removal (middle, PolarNet), and PC with PGR (right). Using PC with semantic ground removal,  detection bounding boxes are mismatched with ground truth (red circle), while PGR does not affect detection performance.
}

    \label{fig:before_after}
\end{figure*}

\section{Introduction}\label{sec:intro}

Point cloud streaming enables various applications, such as augmented reality, sensor fusion, and cooperative perception~\cite{autocast, avr, fcooper}.  For example, in the context of connected and automated vehicles (CAVs), merging data from peer vehicles enhances situation awareness beyond individual on-board sensing capabilities~\cite{v2vnet, cobevt, v2xvit}. 
However, storing and transmitting point clouds from LiDAR (Light Detection and Ranging) sensors is costly.
A point cloud (PC) frame is represented as an array of points, each with ($x$, $y$, $z$) spatial coordinates and associated attributes, such as reflectance.
A 64-line LiDAR generates around 100,000 points per frame, meaning that at 10 FPS scanning rate, the raw data rate of 128 Mbps exceeds the available bandwidth of current vehicle-to-infrastructure (V2I) communication technologies, such as C-V2X \cite{coopernaut}. Therefore, it is necessary to develop efficient systems for vehicular LiDAR data transmission.

The main challenges of designing such efficient systems end-to-end are twofold: 1) a data compression technique that can fit narrow communication bandwidth in which the compressed representation is suitable for remote machine vision tasks, and 2) a low-latency compression pipeline that supports latency-sensitive machine vision tasks. 
To reduce data volume, previous work~\cite{graziosi2020overview} exploits the redundancy in the raw frame to compress the point cloud in a lossless fashion~\cite{schnabel2006octree}, or slightly trades off the fidelity for a lower bit rate~\cite{schwarz2019emerging}.
%
% \todo{talk about two desirable properties here -- first compression machine vision, the second is the latency of such compression... }
%
% Current compression schemes for PCs concentrate on compressing the entire PC. 
However, manipulated data in the compression process, though sometimes indistinguishable by human eyes, can significantly degrade machine vision performance.
To preserve performance, machine vision models such as object detection and semantic segmentation can be use \textit{before} quantization, but this approach only transmits incomplete or encoded information, limiting downstream tasks. In addition, these models often add significant latency, hindering real-time streaming applications.
% \todo{break here and form a new paragraph of in this paper, we explore...}
In this paper, we explore an efficient design that achieves low-latency compression without sacrificing application accuracy.
% In this context, there is potential for further bit rate reduction, since different portions of the PC can have different importance to machine vision systems; in particular, a non-negligible portion of Lidar points are likely ground points. %Specifically, an average of 35 [PERCENT]\% of LiDAR points are ground points. 
Taking object detection as an example, the performance dependency on each point varies significantly. 
\figref{fig:before_after} shows the objects detected by PVT-SSD \cite{yang2023pvt} using the original PC (left), and using the PC after removing ground points (middle) with a semantic segmentation model PolarNet~\cite{zhang2020polarnet}.  It illustrates that naively removing ground points based only on semantics may negatively impact perception performance, as state-of-the-art (SOTA) models show strong dependency on certain ground points. 
Our key idea is to investigate the feasibility of wisely removing irrelevant context (\textit{i.e.} partial ground in this case), while maintaining the performance of downstream computer vision tasks (\figref{fig:before_after} right).

Our main contributions are: (a) A feasibility and innovative study demonstrating that {\it a careful selection} of ground points can be removed with minimal impact on detection accuracy, (b) Experimental results on the value of retaining partial ground points close to objects, and (c) A lightweight, highly parallelizable \textbf{P}illar-based \textbf{G}round point \textbf{R}emoval (PGR) algorithm that is able to selectively retain most ground points near objects without the complexity of detecting objects.

%\begin{itemize}
%    \item A feasibility study demonstrating that {\it most} ground points can be removed with minimal effect on detection accuracy,
%    \item Experimental results on the value of retaining ground points that are close to objects,
%    \item A computationally simple and highly parallelizable pillar-based ground point removal algorithm that is able to selectively retain most ground points near objects without the complexity of detecting objects.
%\end{itemize}
 \section{Related work} \label{sec:related}

%In the following subsections, we review PC compression, ground point removal, and PC object detection.
%In this section, we review the relevant literature regarding the major aspects of our research, giving the necessary background to fully understand our work.
%In section \ref{pcc}, we delve into significant techniques for compressing point clouds, while Sections \ref{gprm} and \ref{ob_det} present state-of-the-arts methods for ground point removal and object detection in point clouds, respectively. 
%Our research is closely related to PC compression, PC ground removal, and PC object detection.

\subsection{Point Cloud Compression} \label{pcc}
A PC
  \(X =\{\boldsymbol{p}_i \in \mathbb{R}^{d}\}_1^N\), 
is a set of $N$ points in 3D space, where each point \(\boldsymbol{p}_i \) consists of spatial coordinates \({x_i, y_i, z_i}\), and additional attributes such as RGB color and reflectance.
Unlike images, this data is unstructured; the points are situated within a vast 3D space characterized by local density but sparse distribution overall.
To efficiently compress a PC, it has to be first converted to a data structure that can represent position data compactly, allowing exploitation of inter-point attribute correlation. 
One approach uses a tree structure \cite{schnabel2006octree, elseberg2013one, schwarz2018emerging, mpeg-g-pcc}, suitable for representing data with unevenly distributed point density. 
An octree-based PC compression algorithm was proposed in \cite{schnabel2006octree}, compactly describes occupancy in 3D space. 
%By partitioning a space with octree structure, a point cloud can be described in compact way and have the flexibility to adapt to different resolution. 
Draco  \cite{draco} uses KD-trees to depict PC geometry. Additionally, MPEG \cite{mpeg-g-pcc} has introduced a standard for encoding and decoding octree-based PCs, enabling compression based on geometry (G-PCC); we exploited this algorithm in our work. 
%With a tree constructed, inter-point attribute correlation can be exploited. Similar to traditional transform-based image and video coding \cite{goyal2001theoretical}, various transforms on attributes have been proposed. The Region-Adaptive Hierarchical Transform (RAHT) \cite{de2016compression} transforms attribute data following the existing octree structure. The transform coefficients are quantized and entropy coded. 

\subsection{Ground Point Removal} \label{gprm}

We categorize Lidar ground point removal methods into non-learning-based methods, which employ handcrafted techniques for ground segmentation, and learning-based methods, which use deep learning.  In the former group, older methods such as \cite{himmelsbach2010fast} partitioned PCs to estimate ground points by comparisons with local line fits, while \cite{moosmann2009segmentation} exploited a  graph-based approach to segment ground and objects based on local convexity measures.
A two-step algorithm \cite{zermas2017fast} identified the ground surface iteratively by using deterministically assigned seed points and clustering the remaining non-ground points, leveraging the structure of the Lidar PC, and \cite{narksri2018slope} detected most non-ground points based on inter-ring distances, then used multi-region RANSAC plane fitting to separate the remaining non-ground. 
More recently, \cite{steinke2023groundgrid} used a recursive algorithm to obtain a 2D elevation map to estimate terrain and segment the PC, while \cite{patchword} encoded a PC into a Concentric Zone Model–based representation, followed by ground plane fitting and ground likelihood estimation to extract the final ground segmentation.
%into a Concentric Zone Model–based representation to assign an
%appropriate density of cloud points among bins, followed by Region-wise
%Ground Plane Fitting and a Ground Likelihood Estimation.

Among learning-based methods, \cite{paigwar2020gndnet} estimated the ground plane elevation end-to-end with a grid-based representation, exploiting PointNet \cite{qi2017pointnet} to extract features and regressing ground height for each cell of the grid.
In \cite{zhang2020polarnet}, a polar bird’s-eye-view representation balanced the points across grid cells in a polar coordinate system and aligned the attention of a segmentation network with the long-tailed distribution of points along the radial axis.
For segmentation, they exploited a simplified KNN-free PointNet to transform points to a fixed-length representation, and then a ring CNN that outputs a quantized prediction, decoded finally to the point domain.

%\textcolor{red}{non-learning based ground segmentation methods:
%\cite{moosmann2009segmentation, himmelsbach2010fast, zermas2017fast, narksri2018slope, steinke2023groundgrid}. recent learning based ground segmentation methods:\cite{paigwar2020gndnet}}
%~\cite{paigwar2020gndnet}

\subsection{Object Detection on Point Clouds} \label{ob_det}
Existing methods for LiDAR 3D object detection can be classified into voxel-based, point-based, and point-voxel based methods. In voxel-based methods~\cite{yang2018pixor, zhou2018voxelnet, yan2018second, lang2019pointpillars, wang2020pillar}, a PC is partitioned into regular spaces called voxels, and features extracted from voxels are fed into deep neural networks. VoxelNet \cite{zhou2018voxelnet} and SECOND \cite{yan2018second} represent seminal works of this approach, while PointPillars~\cite{lang2019pointpillars} further reduces computational complexity by defining pillars that extend only along the vertical axis.
Similarly, PillarNet~\cite{pillarnet} introduces a powerful encoder network for effective pillar feature learning and a neck network for spatial-semantic feature fusion.
Point-based methods directly use the PC's unstructured format without converting it into voxels. The pioneering work PointNet~\cite{qi2017pointnet} was followed by extensive improvements \cite{qi2017pointnet++, shi2019pointrcnn, yang20203dssd, pan20213d, shi2020point}. Point-based methods use point sampling, which aims to select a representative subset of input points, and feature learning, which learns local features related to the selected points, to be used by subsequent object detection layers. 

%Point-voxel-based methods \cite{miao2021pvgnet, shi2020pv, yang2022graph, ye2020hvnet} combine voxel-level and point-level features. PV-RCNN \cite{shi2020pv} combines voxel-based CNN features and PointNet-based features into a set of keypoints during the proposal refinement stage. 
Since the transformer architecture \cite{vaswani2017attention, dosovitskiy2020image} has shown great ability in vision tasks, there have been works with transformers \cite{liu2021group,misra2021end,dvst,yang2023pvt}. PVT-SSD \cite{yang2023pvt} uses a transformer architecture to associate contextual features from voxels and geometric features from points.

% \begin{figure*}
%     \centering
%     \includegraphics[width=0.8\textwidth]{figures/method/intuition_pvt-ssd.png}
%     \caption{Feasibility study on ground point removal.\hangq{Please re-gen this figure asap, remove top caption as said by Pam, To prove the point, maybe a EF=[0, 1.2, 3.6] is sufficient. Also, consider dropping peds and cyclist results if not useful. Finally, perhaps bpp here is not relevant, if all you need to say is removing all grounds is not good, pick one scaling factor for all EF should be enough. The full results can be moved to appendix or something}}
%     \label{fig:intuition_pvt-ssd}
% \end{figure*}

\begin{figure}
    \centering
    \includegraphics[width=0.8\columnwidth]{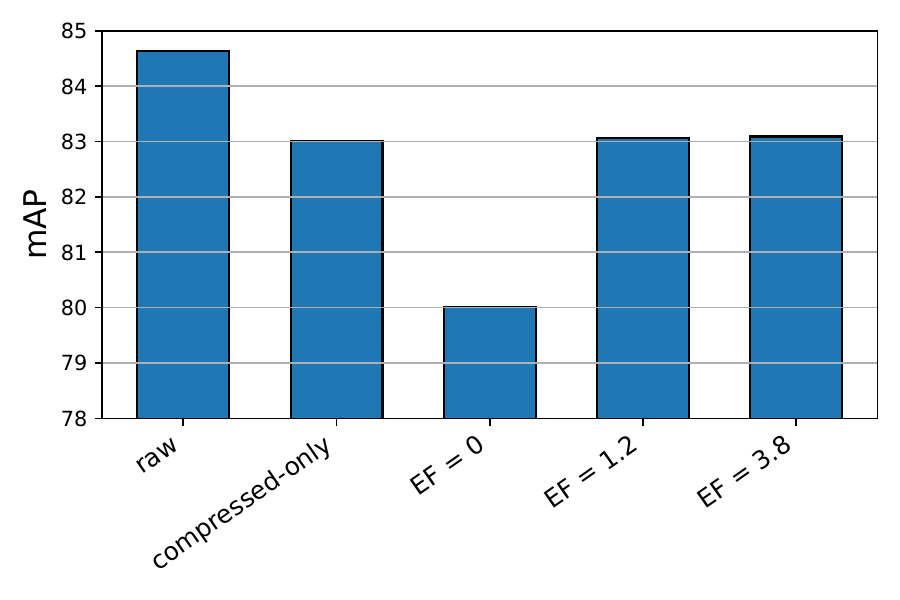} %intuition_pvt_ssd_kitti_car.pdf 
    \vspace{-0.25cm}
    \caption{Feasibility study on ground point removal for Car  detection in  KITTI Dataset. The evaluation ranges from uncompressed 'raw' PCs to 'compressed-only' without ground point removal, and then through a series of Extension Factors (EFs) combined with compression.}
    \label{fig:intuition_pvt-ssd}
\end{figure}

\section{Methods} \label{sec:methods}

%In this section we detail the novel method we have developed to  optimally remove the ground points from a PC. First, in sec. \ref{problem_statement} we formalize the problem and our goal, then  in \ref{impact_rgp} we analyze the impact of removing ground points on object detection, justifying thus the necessity of our algorithm. In sec. \ref{Pillar_based_GR} we describe our method, which is composed by two stages, i.e.  \emph{Pillar Removal} and \emph{Pillar Restoration}. Finally, in sec. \ref{Point_cloud_compression} we illustrate how we performed PC compression exploiting G-PCC \cite{mpeg-g-pcc}   

\subsection{Problem Statement} \label{problem_statement}
Given an input PC frame \(X =\{\boldsymbol{p}_i \in \mathbb{R}^{d}\}_1^N\), a ground point removal pre-processing algorithm \(f\), a PC encoder \(g_{\text{enc}}\), and a PC decoder \(g_{dec}\), the encoded bitstream is
$   b = g_{\text{enc}}(f(X)) $
while the decoded PC is represented as
$   Y = g_{\text{dec}}(b).$
Considering a  downstream machine vision task T with its corresponding performance denoted \(p_{T}(Y)\), the goal of this work is to find a suitable $f$ that allows a significant bit rate reduction with a negligible sacrifice in terms of \(p_{T}(Y)\).
We focus only on the choice of $f$, and we maintain all other parts of the process fixed. 

%Consider an input PC frame \(X =\{\boldsymbol{p}_i \in \mathbb{R}^{d}\}_i^N\), a PC pre-processing algorithm \(f\), a point cloud encoder \(g_{\text{enc}}\), and a point cloud decoder \(g_{dec}\). The encoded bitstream is: $$b = g_{\text{enc}}(f(X))$$
%The decoded point cloud is:
%$$Y = g_{\text{dec}}(b)$$
%For a downstream machine vision task \(T\), the performance is denoted \(p_{T}(Y)\).
%For fixed standard encoder \(g_{enc}\) and decoder \(g_{dec}\), we want to find \(f\) that results in a small sacrifice in \(p_{T}(Y)\) with significant bitrate reduction. 

\subsection{Impact of Ground Points on 3D Object Detection} \label{impact_rgp}
In this part we examine how the total or partial removal of ground points impacts the performance of the downstream machine vision task.
Starting from an input PC $X$, we consider a pre-processing algorithm \(f(X)=\hat{X}\) that precisely removes selected ground points.
We investigate the feasibility of this approach in an omniscient way, which means that we use the SOTA method PolarNet \cite{zhang2020polarnet} to extract the ground points in a PC; the points labeled as ground in the segmentation result are used as a reference for ground point removal experiments.

% \begin{figure*}[htb]
%     \centering
%     \includegraphics[width=0.8\textwidth]{figures/method/intuition_pvt-ssd.png}
%     \caption{Feasibility study on ground point removal.}
%     \label{fig:intuition_pvt-ssd}
% \end{figure*}

In the experiments, we first removed all ground points and then restored those in the extended bounding boxes of objects annotated in ground-truth labels. The system can be described as:
\begin{align}
    & \hat{X}=f(X;S,EF) \\
    & Y=g_{dec}(g_{enc}(\hat{X}))
\end{align}
%$$\hat{X}=f(X;S,EF)$$
%$$Y=g_{dec}(g_{enc}(\hat{X}))$$
where \(S\) is the per-point semantic segmentation result from PolarNet,  and $EF$ is an extension factor which controls how far we extend from the ground truth object bounding boxes. For example, if the ground truth bounding box for a car has the size $(length, width, height)$, then using EF, we restore any ground points extracted by PolarNet in the box of size $(1+EF)\times(length, width, height)$. When \(EF=0\), it means that no extension is applied and only the annotated object points are restored.
The PC after pre-processing will be compressed and decompressed with the standard geometry-based PC compression method G-PCC~\cite{mpeg-g-pcc}. The decompressed point cloud \(Y\) is used for downstream task performance evaluation. 

We exploit PVT-SSD \cite{yang2023pvt} to study the performance degradation introduced by ground point removal. 
We compare omniscient point removal with a baseline where the raw PC is encoded without any point removed (\(\hat{X}=X\)). Using different extension factors, we found (\figref{fig:intuition_pvt-ssd}) that for the Car class, removing all ground points degrades detection accuracy by more than 3\%. However, with \(EF\in[0.3, 3.6]\), the partial removal of ground points allows the system to achieve the same detection accuracy as for a compressed PC with no ground points removed, but at significantly lower bit rates.  This indicates that {\it while full removal of the ground points outside of the object bounding boxes hurts detection, partial removal holds potential for rate savings without sacrificing detection}.
%operating points with bit rates greater than 2.0bpp have detection accuracy close to the no-preprocessing baseline. 
%It is noteworthy that with \(EF=0.3\), the detection accuracy is already significantly improve than \(EF=0\).This means the ground point removal can extend more acceptable rate-accuracy operating points to lower bit rates, while maintaining performance close to no-preprocessing baseline. 

\begin{figure}
    \centering
    \includegraphics[width=0.8\columnwidth]{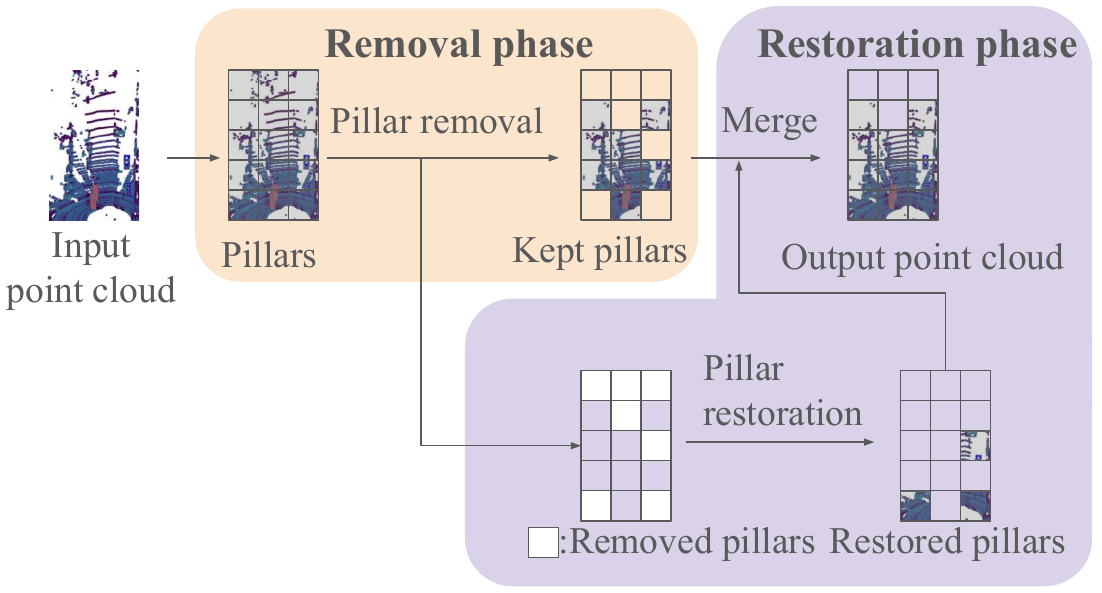}
    \caption{Illustration of PGR with reduced number of pillars.}
    \label{fig:pillarization}
\end{figure}

\subsection{Pillar-based Ground Removal Algorithm} \label{Pillar_based_GR}
The observation from the feasibility study gives us intuition on designing a pillar-based ground removal (PGR) algorithm  that removes ground points that are not near objects.
As shown in \figref{fig:pillarization}, our method has two steps, \textbf{Pillar removal} and \textbf{Pillar restoration}.
%The algorithm has two phases, i.e.  removal and reconstruction, as shown in \cref{fig:pillarization}

%{\it Pillar Removal:}
\subsubsection{Pillar removal}
%This first phase removes points most likely to be ground points.
Starting from the assumption that in a sufficiently small region composed of only ground points, there would be little difference in height,
every input frame is split into a 2D grid of square pillars $pl_{j}$ for $j = \{1,...,M \}$, where $M$ is the total numbers of pillars. 
The pillar size, or \emph{resolution}, determines the granularity at which the overall algorithm operates.
The goal is to remove pillars that are likely part of the ground.
For each pillar, if the height difference between its highest and lowest points is below a certain threshold, then it might contain only ground points.
For a pillar $pl_i$, this condition is written as:
\begin{equation} \label{cond_1}
    d_z(pl_i)  = z_{max}(pl_i) - z_{min}(pl_i) \leq \delta_{minmax} 
\end{equation}
where $z_{max}(pl_i)$ and $z_{min}(pl_i)$ are the maximum and minimum heights in the pillar, and $\delta_{minmax}$ is a threshold. 

However, this condition is insufficient-- if for example, a pillar passes through the roof of a car, it is possible that the height difference is small because this area is quite flat; erroneously detecting these points as ground would mistakenly remove a portion of the car.
In addition to the condition on height difference, one needs to consider the neighboring area and a \emph{local ground baseline}.
We compare $z_{min}(pl_i)$ with the height $b$ of the lowest point in a square neighborhood controlled by a parameter \emph{environmental radius} (\emph{er}), representing half of the side length.  A pillar is considered to be ground if it fulfills condition \eqref{cond_1} and also the following one:
\begin{equation}
    d_{env}(pl_i) = z_{min}(pl_i) - b < \delta_{env}.
\end{equation}
Summing up, the pillar removal step consists of an indicator function $\Phi_{i}$:

\begin{equation}
    \Phi_{i} =
  \begin{cases}
    0       & \quad \text{if }  d_z(pl_i)  \leq \delta_{minmax}  \text{ and } d_{env}(pl_i) < \delta_{env}   \\
    1  & \quad \text{otherwise} 
  \end{cases}
\end{equation}

% {\textbf{Parameter tuning:}}
Parameter choices should generally be based on physical aspects of PCs rather than the nature of a specific dataset.
Concerning  $\delta_{minmax}$,  it should be large enough to allow sidewalk curbs (typically 10-20cm) as well as somewhat taller street medians to be considered ground points and removed, and yet small enough to avoid deleting small pedestrians (typical 2-year-old children are 80-95cm tall);  
So we set $\delta_{minmax}$ to 40cm.
We set \emph{er} to 1.8m, roughly the width of a standard car, aiming to ensure that a pillar containing only a flat car roof will get compared with the local ground baseline and not be declared as ground points.
We set $\delta_{env}$ to 40cm, to ensure that a flat car roof or even the top of a flat baby stroller is differentiated from the local ground baseline.
We emphasize that these values are based on physical characteristics of the environment rather than on individual datasets, which should enhance the algorithm's robustness. Lastly, a smaller \emph{resolution} means that we consider smaller pillars. The choice trades off grid granularity and model speed; empirically we set it to 40cm. \\

\subsubsection{Pillar restoration}
%{\it Pillar Restoration:}
The \emph{Pillar removal} stage can remove points close to an object; however, we observed in Sec.~\ref{impact_rgp} that we need ground points in the vicinity of objects for better detection precision. Operating on the pillar level, the \emph{Pillar restoration} step restores some previously removed pillars.
If a removed pillar $pl_{i}$ is close enough to at least one retained pillar, then preserving $pl_{i}$ is likely to be useful for the machine vision task.
A removed pillar $pl_{i}$ is restored if there exists any pillar $pl_{j}$ retained during the removal phase such that the Chessboard distance between their centers is below a threshold $\delta_{res}$.
We use $R_i$ to denote a restoration flag indicating whether to restore $pl_i$:
\begin{equation}
\label{reconstruct_pillar}
   R_{i} =
  \begin{cases}
    1      & \quad \text{if } \sum_{j | D(i,j)\leq \delta_{res}} \Phi_{j} >0   \\
    0  & \quad \text{otherwise.} 
  \end{cases}
\end{equation}

% {\textbf{Parameter tuning:}}
As points become sparser farther from the Lidar, objects may lose critical points that contain information about their global geometry.
Such important points could be close to the ground and thus removed in the removal phase, but cannot be restored if $\delta_{res}$ is not large enough.
To strike a good trade-off between removal efficiency and detection precision, we adapt $\delta_{res}$ for different ranges.
From the experiments, for pillars closer than 30m, we used $\delta_{res, 1}=1.8\text{m}$ for KITTI and $\delta_{res, 1}=2.2\text{m}$ for the Waymo Open Dataset.  For both datasets, we used $\delta_{res, 2}=5.4 \text{m}$ for pillars further away. Due to the sparsity of points in far ranges, a large $\delta_{res}$ will cause only a slight increase in the number of ground points remaining.

Fig.~\ref{fig:pillarization} depicts the preprocessing framework, which is
is not recursive and is applied only once, obtaining real-time runtime.

\begin{figure*}[!t]
\centering
\begin{subfigure}{0.7\linewidth}{\includegraphics[width=\linewidth]{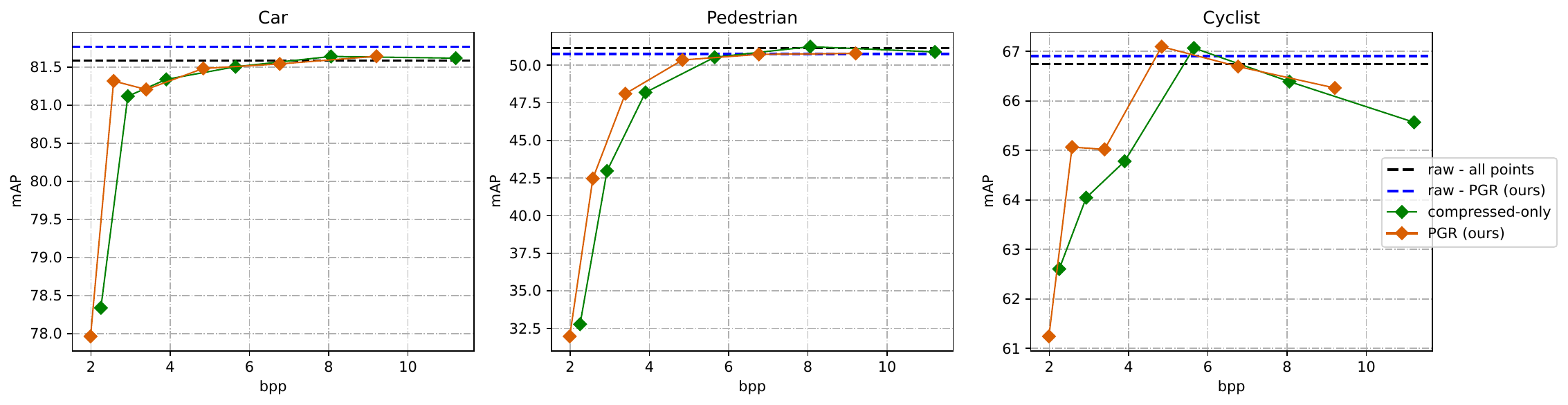}}
\vspace{-0.5cm}
\caption{bpp vs mAP using SECOND.}
\label{fig:second_kitti_precision_vs_bpp}
\end{subfigure}
\begin{subfigure}{0.7\linewidth}{\includegraphics[width=\linewidth]{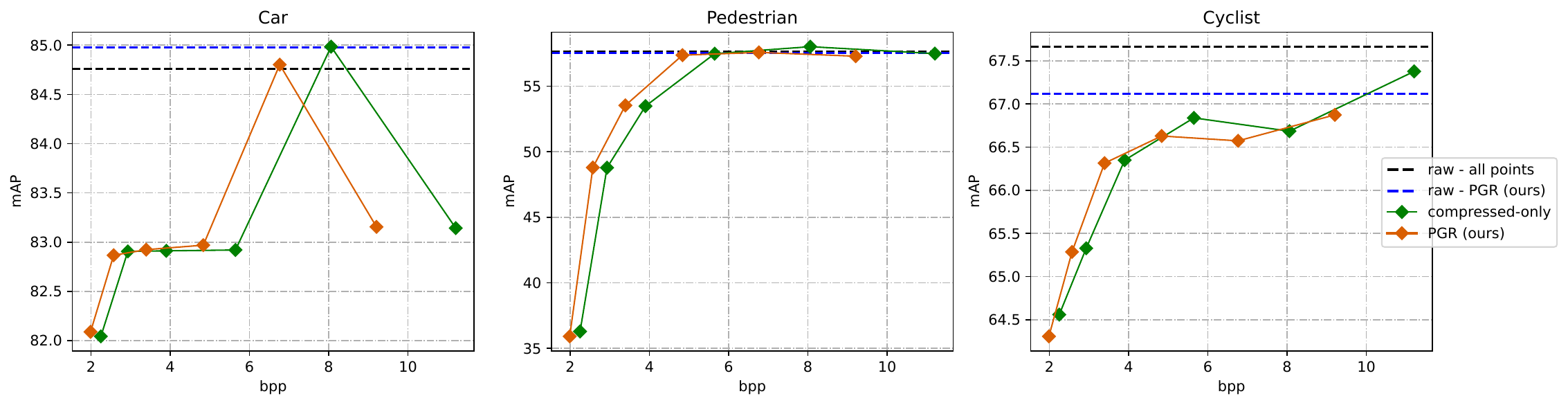}}
\vspace{-0.25cm}
\caption{bpp vs mAP using PVT-SSD.}
\label{fig:pvt_ssd_kitti_precision_vs_bpp}
\end{subfigure}

\caption{bpp vs mAP  on KITTI dataset using (a) SECOND and (b) PVT-SSD. Dotted lines represent results on uncompressed data. The up arrow $\uparrow$ after mAP means that higher is better and the down arrow $\downarrow$ after bpp means that lower is better.}
\label{fig:kitti_precision_vs_bpp}
\end{figure*}

\section{Experiments}\label{sec:evaluation}

\subsection{ Experimental Setup}
We selected six configurations of geometry scaling factors
(0.01, 0.012, 0.015, 0.022, 0.035, 0.063) paired up with six attribute quantization parameters (34, 30, 26, 22, 18, 14).
The bit rate of the compressed PC is represented by bits per point (bpp), obtained by dividing the total number of bits in the compressed frame representation by the total number of points in the PC for that frame.

We use object detection to evaluate the effectiveness of PGR. 
The PC after applying PGR is compressed and decompressed and subsequently sent through pre-trained object detection models. 
We used \textbf{Mean Average Precision (mAP)} as the detection performance metric for each class
%, which is based on the ranking of detection scores for each class  
\cite{map}.

\subsection{Datasets}
\subsubsection{KITTI} The KITTI dataset \cite{geiger2012we} contains 7481 training samples and 7518 test samples. The training samples are commonly divided into a training set with 3712 samples and a validation set with 3769 samples. Annotations are provided for objects in the camera's field of view. Objects are labeled into classes including Car, Pedestrian, and Cyclist. Depending on three factors (minimum bounding box size, occlusion level, and maximum truncation), the annotated objects are classified into easy, moderate, and hard difficulty levels. Common practice is to report performance metrics on moderate difficulty.

\subsubsection{Waymo} The Waymo Open Dataset (WOD) \cite{Sun_2020_CVPR} contains 798 scenes for training and 202 scenes for validation. Annotations are provided separately for Lidar and camera data. Based on the number of points contained, objects are classified into two difficulty levels, LEVEL\_1 for those having more than five points, and LEVEL\_2 for those having at least one point. Each Lidar data entry consists of the point's (x, y, z) coordinates and attributes (intensity and elongation). Each frame may contain a {\it no-label zone} in which there are no ground truth annotations attempted, usually because the points are far from the sensor and difficult to annotate or considered unimportant. In this paper, we remove the points in the no-label zone since, without ground truth labels, these regions would not contribute meaningfully to the detection performance.
WOD frames are in the format of a range image where pixel values represent the distance between the point and LiDAR, and two attributes, intensity and elongation, are given. In the experiments, we transformed the range image format to 3D format where each entry contains a point's coordinates and attributes in 32-bit float.

\begin{comment}
\begin{figure*}[tb]
 \begin{subfigure}{0.98\linewidth}{\includegraphics[width=\linewidth]{figures/evaluation/second_waymo.pdf}}
\caption{mAP vs bpp using pre-trained SECOND model.}
\label{fig:second_waymo_precision_vs_bpp}
\end{subfigure}
\begin{subfigure}{0.98\linewidth}{\includegraphics[width=\linewidth]{figures/evaluation/pvt_ssd_waymo.pdf}}
\caption{mAP vs bpp using pre-trained PVT-SSD model.}
\label{fig:pvt_ssd_waymo_precision_vs_bpp}
\end{subfigure}

\caption{Object detection performance (mAP) versus bit per point (bpp) on Waymo Open Dataset using (a) SECOND and (b) PVT-SSD. Dotted lines represent results on uncompressed data.}
\label{fig:waymo_precision_vs_bpp}
\end{figure*}   
\end{comment}

\begin{figure*}[tb]
\centering
    \includegraphics[width=0.7\textwidth]{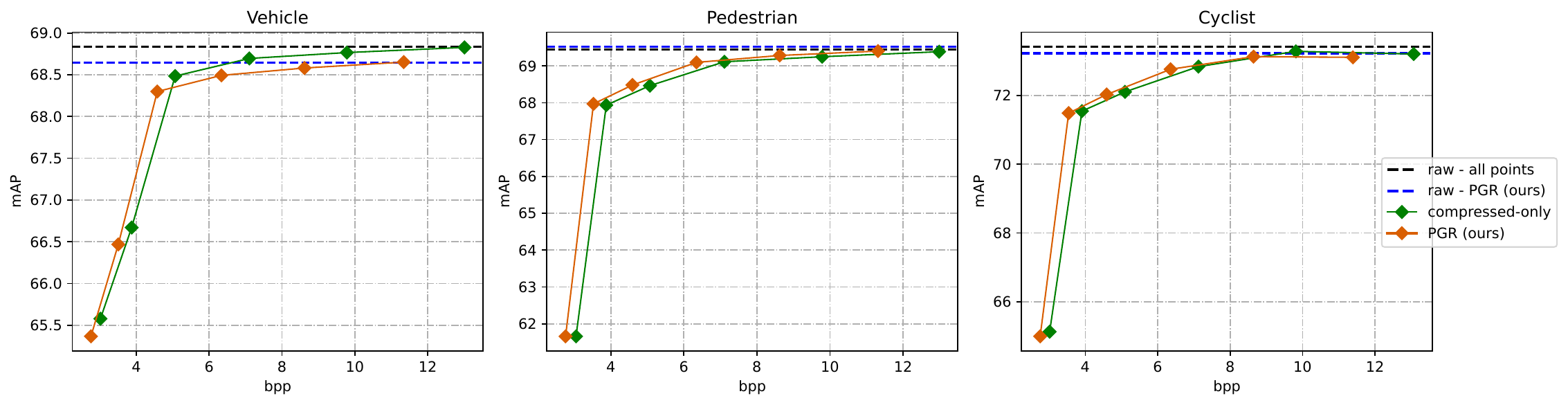}
  \vspace{-0.15cm}
\caption{bpp vs mAP  on Waymo Open Dataset using PVT-SSD. Dotted lines represent results on uncompressed data.}
\label{fig:pvt_ssd_waymo_precision_vs_bpp}
\end{figure*}

\subsection{\ourname  Performance on Object Detection} \label{results}
In this section, we evaluate the effectiveness of \ourname as a preprocessing function before encoding the PC, considering  two different object detection models, SECOND \cite{yan2018second} and PVT-SSD \cite{yang2023pvt}.
%and we consider 6 different levels of compression, covering a bit range of about $\sim[ 2,12]$ for Waymo and  $\sim [ 2,10]$ for KITTI.

%\input{figlatex/ablation_fig}

\subsubsection{bpp vs. mAP}
Figures \ref{fig:kitti_precision_vs_bpp} and \ref{fig:pvt_ssd_waymo_precision_vs_bpp} show the bpp vs. MaP for the two datasets. 
Solid lines depict results with data compressed at various levels, with orange for \ourname and green for the entire PC (no ground points removed).  Dashed lines represent results on the raw dataset (no compression).
%, either preprocessed with \ourname (blue) or not (black).
%For each plot, the $i$-th point from the left on the green curve is generated by the same compression configuration as the $i$-th point from the left on the orange curve;  by comparing them, we can see \ourname generally reduces the compressed bit rate.
Although the rate reduction depends on the operating point, \ourname effectively shifts the rate-performance curve to the left, generally reducing the bitrate;  this means we can achieve the same desired task performance with fewer bits. 
In~\figref{fig:kitti_precision_vs_bpp}, the \ourname curve is above the non-processed curve for most of the range, for both models and all objects classes.
For Waymo, there are gains for the Pedestrian and Cyclist classes, while for the Vehicle class, our curve is slightly below the compressed-only baseline. However, at lower bpp we can still observe an improvement, suggesting it may be of value to apply \ourname in the low bit rate regime.
The fraction of points that \ourname removes depends on the frame, ranging from about $10\%$ in dense urban areas to more than $50\%$ in open ground. 

Tab. \ref{tab:keep_rates} shows the average percentage of points that \ourname preserves; it manages to remove a large portion of ground points with very little removal of relevant objects; in~\figref{fig:pvt_ssd_waymo_precision_vs_bpp}, comparing the lowest bit rate configurations, \ourname achieves a 9.34\% bit rate savings (0.282 bpp) with minimal mAP changes: -0.209\% (Vehicle), -0.003\% (Pedestrian), and -0.154\% (Cyclist) with respect to no GP removal.
For the highest bit rate configurations, bit rate reduces by 12.94\% (1.685 bpp) with mAP changes of -0.175\%, +0.011\%, and -0.078\% for the respective classes.

\subsubsection{Run-time}
Algorithm complexity is important if ground point pre-processing were to be done in real time.
%PGR has a significant advantage in running speed. The operations for every pillar are fully parallelized. 
On a GeForce RTX 3090 with 8 CPU cores, the speed to process frames sequentially (one by one), can reach 86.5 frames per second (fps), 11.6 ms per frame.
% so can be considered real-time. 
This leaves ample latency budget for 
% further processing 
 downstream tasks such
as object detection, prediction, and segmentation.

\begin{table}[h]
  \caption{Avg Percentage (\%) of points preserved by \ourname}
\vspace{-0.25cm}
  \label{tab:keep_rates}
  \begin{center}
    \label{tab:table1}
    \begin{tabularx}{0.70\columnwidth}{l X X} % <-- Changed to S here.
    \toprule
      \textbf{Points Category} & \textbf{KITTI} & \textbf{Waymo}\\
      %\hline \\[.05cm]
      \midrule
      Car (KITTI) / Vehicle (WOD) & 99.981  & 99.980 \\
      Pedestrian &  99.995 & 99.982 \\
      Cyclist &  99.995 &  99.988\\
      Preserved points & 75.133 &  82.415\\
      \bottomrule
    \end{tabularx}
  \end{center}

\end{table}

\begin{figure*}[ht]
\centering
    \includegraphics[width=0.75\textwidth]{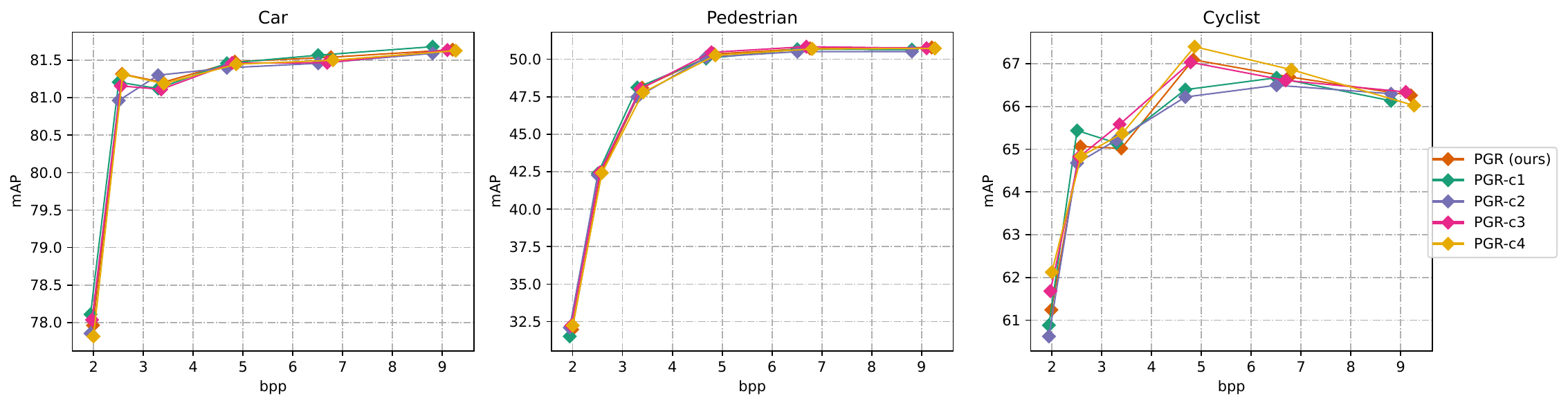}
\vspace{-0.15cm}
\caption{bpp vs mAP on KITTI Dataset using SECOND, considering \ourname with different configuration}
\label{rob}
\end{figure*}

\begin{table}[!t]
\caption{ BD-mAP of our \ourname with and without restoration phase, against SOTA method for point ground removal used as pre-processing and PVT-SSD as pre-trained model. We consider the case \emph{compressed-only} as anchor.} 

 \label{tab:ab}
%\small
\centering
\vspace{-0.25cm}
\resizebox{0.5\textwidth}{!}{
\begin{tabular}{l   l    | l    | l     }
\toprule
 \multicolumn{1}{c}{}  & \multicolumn{1}{c|}{\emph{Car}}  & \multicolumn{1}{c|}{Pedestrian} & \multicolumn{1}{c}{\textbf{Cyclist}} \\
\cmidrule(l){2-2} \cmidrule(l){3-3} \cmidrule(l){4-4}
 &    BD-mAP   & 
 BD-mAP        & 
 BD-mAP    \\
%\cmidrule(l){2-4} \cmidrule(l){5-7}  \cmidrule(l){8-10}
\midrule
%\emph{Compressed-only} & 000   & 000  & 000  &000 & 000    & 000   & 000 & 000 & 000  \\
PGR (ours)  & \textbf{-0.14}  &1.75   & \textbf{0.008}    \\
PGR (ours) w/o restoration  & -1.89   & \textbf{4.36}    & -0.57   \\
GnDNet   & -4.16   &-13.72    & -6.388    \\
PolarNet   & -1.42    &-0.427     & -0.95    \\
patchwork++  & -1.38    &-0.97    & -0.949  \\
\bottomrule
\end{tabular} \label{tab_training_cost}
}
\end{table}

\subsection{Robustness  with respect to parameter tuning}
Robustness to parameter tuning is crucial for this type of system.
Here, we evaluate the performance when varying the parameters introduced in Sect.~\ref{sec:methods}. The parameters have specific real-world physical meanings that remain consistent across datasets (KITTI and Waymo) and object detection models.
Fig.~\ref{rob} shows the results of mAP vs. bpp on the KITTI dataset using SECOND, considering different configurations of our method.
The orange line (PGR-c0) represents the standard configuration introduced in Sect. \ref{sec:methods}, while in the other scenarios we manually changed parameters as follows:

\begin{itemize}
    \item PGR-c1: \emph{er} passes from 1.8 to 1.4.
    \item PGR-c2: $\delta_{res, 1}$ passes from 1.8 to 1.4.
    \item PGR-c3:  $\delta_{minmax}$ = 0.6.
    \item PGR-c4: (\emph{er}, $\delta_{minmax}$,$\delta_{res, 1}$,$\delta_{res, 2}$) = (0.6,0.35,1.6,5.2) 
\end{itemize}

Here we mention only the values that have been changed. The mAP remains similar in all configurations; for both \emph{car} and \emph{pedestrian} the differences are negligible, with only modest variations for the cyclist in the central bit range, without catastrophic degradation. Even for PGR-c4, where we modified 4 parameters, the final performance does not change remarkably, demonstrating the robustness of \ourname to parameter tuning.
Notably, the configurations for KITTI and Waymo in Sec. \ref{sec:methods} are essentially the same, showing consistent performance in all datasets without significant parameter adjustments.

\subsection{Comparison with other Ground Point Removal Methods} \label{comparison}
In this section, we compare \ourname with other ground removal method used for pre-processing, namely PolarNet \cite{zhang2020polarnet}, GndNet \cite{paigwar2020gndnet}, (deep-based methods) and patchwork++ \cite{patchword} (handcrafted method). 
Table~\ref{tab:ab} shows results for KITTI using PVT-SSD as the pretrained model for object detection.
We exploited Bjontegaard \cite{bjonte} metric considering mAP, recalling that a larger BD-mAP  indicates better encoding efficiency. Although in the object detection scenario BD-mAP is not commonly used, they are well suited to summarize results across different compression levels.

\begin{figure}
    \centering
    \includegraphics[width=0.60\columnwidth]{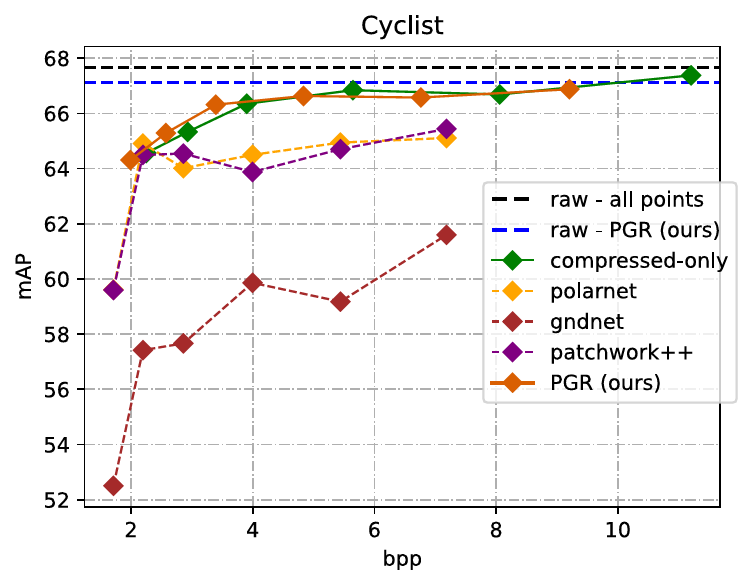} 
\vspace{-0.15cm}
    \caption{bpp vs mAP of \ourname against other ground point removal method considering Cyclist class. }
    \label{fig:cyclist}
\end{figure}

\ourname outperforms all models across all object classes, matching or exceeding the performance of the compressed-only anchor, where no ground points are removed. Fig. \ref{fig:cyclist} further illustrates these results for the Cyclist class.

These results are consistent with Sec.~\ref{impact_rgp}, where we show that removing all ground points leads to a marked performance degradation. While other methods aim to completely remove ground points, \ourname preserves object detection performance with a careful selection of what part of the ground could be removed harmlessly. Because of this, even if \ourname may perform poorly in ground detection compared to other methods (75\% of ground points are preserved in KITTI), we obtained the best results in terms of object detection. The goal of this work is to remove only that portion of the ground that does not degrade performance.

\begin{comment}
    %  figures/evaluation/clist.pdf
\end{comment}

\subsection{Ablation Study} \label{ablation}

Table~\ref{tab:ab} shows results with and without the \ourname restoration phase.
This phase aims to restore some points that were mistakenly removed in the algorithm's first step.
Despite an obvious slight gain in terms of bits saved,  we observe a significant decrease in precision, especially for the Car and Cyclist classes. 
Without the restoration phase, we retain 98. 076\% of car points, 98. 584\% of cyclist points, and 98. 844\% of pedestrian points, values closely matching those in Tab. \ref{tab:keep_rates} with restoration; the latter class is the only one that presents some degradation after the restoration phase.
This suggests that for smaller objects, detection models may require fewer ground points around them to provide context, and that with a more compact geometry, critical points are less likely to be lost during the removal phase.

\begin{comment}

\subsection{Run-time} \label{speed}

Algorithm complexity is important if ground point pre-processing were to be done in real time.
%PGR has a significant advantage in running speed. The operations for every pillar are fully parallelized. 
On a GeForce RTX 3090 with 8 CPU cores, the speed to process frames sequentially (one by one) can reach 86.5 frames per second (fps), 11.6 ms per frame.
% so can be considered real-time. 
This leaves ample latency budget for 
% further processing 
 downstream tasks such
as object detection, prediction, and segmentation.

\end{comment}

\section{Conclusion and future works}
\vspace{-0.04cm}
This work provides insight that some ground points are needed in PCs to enable remote detection of cars, pedestrians and cyclists.  We then present a novel method to remove superfluous ground points.
Following the intuition shown in Section \ref{impact_rgp} that portions of the ground close to objects could be useful for downstream machine vision tasks, we devised a two-step algorithm for targeted ground removal.
First, we remove points that are most likely part of the ground, and then restore those points that are close enough to objects.
Although simple, \ourname produced excellent results in bit rate reduction, without compromising the final precision results for object detection, especially in low bit rate regimes, emphasizing also that \ourname can be considered real-time, achieving a speed of 86.5 fps.
We also showed in Sec.~\ref{comparison} that our method is more suitable than other SOTA methods for this preprocessing, yielding better results for all objects considered, analyzing also in Sec. \ref{ablation} the impact of the restoration phase.

Future developments will focus on enhancing the algorithm's robustness to uneven surfaces and small objects, and optimizing parameter selection using Bayesian optimization or evolutionary algorithms.

%%%%%%%%%%%%%%%%%%%%%%%%%%%%%%%%%%%%%%%%%%%%%%%%%%%%%%%%%%%%%%%%%%%%%%%%%%%%%%%%

\end{document}